\documentclass[letterpaper,10pt,twocolumn]{article}
\usepackage[letterpaper,left=0.75in,right=0.75in,top=0.75in,bottom=1.0in,
            columnsep=0.25in]{geometry}
\usepackage{times}
\usepackage[T1]{fontenc}
\usepackage{titlesec}
\titleformat{\section}{\normalfont\large\bfseries}{\thesection}{0.6em}{}
\titleformat{\subsection}{\normalfont\normalsize\bfseries}{\thesubsection}{0.6em}{}
\titlespacing*{\section}{0pt}{1.2ex plus .4ex}{0.7ex}
\titlespacing*{\subsection}{0pt}{1.0ex plus .3ex}{0.5ex}
\usepackage[utf8]{inputenc}
\usepackage{graphicx}
\usepackage{booktabs}
\usepackage{amsmath,amssymb}
\usepackage{caption}
\usepackage{algorithm}
\usepackage{algpseudocode}
\usepackage{xcolor}
\usepackage{url}
\usepackage{authblk}
\usepackage[hidelinks]{hyperref}

\graphicspath{{./}}

\newcommand{\iso}{\textsc{isolated}}
\newcommand{\mix}{\textsc{mixing}}

\title{\bf What Makes a Redundant Representation Remember?\\
       Lineage Isolation, Not Masking}
\author{
  Jia Huang\textsuperscript{1}, Yangjun Ou\textsuperscript{2}
}
\affil{
  \textsuperscript{1}Guanghua School of Management,Peking University\\
  \textsuperscript{2}School of Mathematical Sciences,Peking University
}
\date{}

\begin{document}
\maketitle

\begin{abstract}
\noindent
Memory-based evolutionary algorithms for dynamic optimization often carry a
redundant second copy of the genotype and expose only one copy to the
objective, on the assumption that the shielded copy accumulates information
about past optima. We show this assumption is false as usually implemented,
and identify the structural property that actually determines whether the
shielded copy retains information. We formalize such methods as a
\emph{gated dual-copy representation} with two independent design axes: a
\textbf{gating rule} deciding which copy is evaluated, and an
\textbf{inheritance rule} deciding whether the two copies mix across
generations. A $2\times2$ ablation shows retained information is governed
almost entirely by the inheritance rule (21.4 vs.\ 1.3 bits) and is nearly
invariant to the gating rule. Per-locus independent inheritance reshuffles
cross-locus structure every generation, so shielding preserves the
\emph{variance} of the hidden copy while destroying the \emph{pattern} that
constitutes a memory. Under isolated inheritance the memory effect is real:
against a single-copy baseline matched for representation budget, the method
gains $+0.010$ AUC when optima recur periodically and loses $0.078$ when they
drift unidirectionally---a $0.089$ separation under otherwise identical
settings, which excludes explanations based on added capacity. We show the
readout rate is also the corruption rate, predicting and confirming an
interior optimum replicated across two implementations. We report one negative
result with a mechanism: dual-copy representations \emph{lower} the mutational
error threshold, because gated expression is a selector rather than a joint
decoder and therefore provides no coding gain. Finally, we document a
benchmarking hazard: on dynamic benchmarks the choice of recombination
operator alone shifted our baseline by $0.062$ AUC, six times the effect size
under study.
\end{abstract}

\section{Introduction}

Dynamic optimization problems require an algorithm to track a moving optimum,
and a recurring design idea is to equip the algorithm with \emph{memory} so
that previously found solutions can be recovered when conditions repeat.
Explicit memory schemes store solutions in an archive. Implicit memory schemes
instead enlarge the representation: each individual carries two copies of the
genotype, only one of which is evaluated, on the reasoning that the
unevaluated copy is shielded from selection pressure and can therefore retain
information about earlier optima. This idea originates in a biological
analogy---recessive alleles persist because selection acts on expressed
phenotypes---and has been implemented in evolutionary computation for four
decades under the names \emph{diploidy} and \emph{dominance}
\cite{BAGLEY,HOLLSTIEN,HOLLAND,GOLDBERG-SMITH,NG-WONG}.

The reasoning contains an untested step. Shielding a copy from
\emph{evaluation} protects the marginal statistics of that copy. But a memory
of a past optimum is not a marginal statistic; it is a \textbf{joint pattern
across loci}. Nothing in the shielding argument protects joint structure from
\textbf{recombination}, which acts on exactly that. A population can retain
every unit of hidden variation while retaining no information about how those
units were once combined.

This paper isolates and tests that step. We find it fails under standard
per-locus inheritance, and we identify the design property that determines
success.

\subsection{Contributions}
\begin{enumerate}\itemsep2pt
\item \textbf{A formalization that separates two conflated design axes.}
      We define a gated dual-copy representation parameterized by a gating
      rule and an inheritance rule (\S3), and show these have been coupled in
      prior implementations.
\item \textbf{A separating $2\times2$ ablation.} Retained information is
      determined by the inheritance rule ($16\times$ difference) and is nearly
      invariant to the gating rule (\S5.1). The mechanism credited in prior
      work is not the mechanism responsible.
\item \textbf{A controlled falsification test.} Comparing periodic against
      unidirectional environments under otherwise identical settings isolates
      the memory contribution at $0.089$ AUC and rules out capacity-based
      explanations (\S5.2).
\item \textbf{A design principle with an empirical optimum.} The gate-flip
      rate is simultaneously the readout rate and the corruption rate,
      predicting an interior optimum; we confirm it and replicate it across
      two implementations (\S5.3).
\item \textbf{A negative result with a mechanism.} Dual-copy representations
      lower the mutational error threshold; gated expression is a selector,
      not a joint decoder, so redundancy yields no coding gain (\S5.4).
\item \textbf{A benchmarking hazard.} Recombination-operator choice shifted
      the baseline by $0.062$ AUC, exceeding the studied effect sixfold
      (\S5.5).
\end{enumerate}

Code and all raw results are released.

\section{Related Work}

\paragraph{Implicit memory in dynamic evolutionary optimization.}
Bagley \cite{BAGLEY} introduced a dual-copy representation with an evolving
gate map, later refined by Hollstien \cite{HOLLSTIEN} and Holland
\cite{HOLLAND} into a scheme where the gating value is attached to and
inherited with each allele. Goldberg and Smith \cite{GOLDBERG-SMITH} evaluated
this on a non-stationary 0-1 knapsack problem with a time-varying constraint
and reported a substantial advantage over a single-copy representation,
attributing it to shielded alleles acting as a distributed probabilistic
memory of recurring states \cite{SMITH-GOLDBERG}. Later work proposed
alternative gate-update mechanisms \cite{NG-WONG,LEWIS-HART-RITCHIE} and
dualism-based schemes \cite{YANG-YAO}. None of these separate gating from
inheritance, and none measure retained information directly. Our results
reproduce the reported advantage but re-attribute it.

\paragraph{Explicit memory and archive methods.}
Archive-based schemes store and reinject past solutions. Relative to these,
the appeal of an implicit scheme is that its stored content is \emph{live}---it
continues to be updated under weak selection rather than being frozen. Our
results show this property is contingent on the inheritance rule, not on
shielding.

\paragraph{Memory in non-stationary learning.}
Fast--slow weight schemes and replay buffers in continual learning address a
structurally similar problem \cite{ER,DER,CLS-ER}. The variance/information
distinction we draw in \S5.1 applies to any method that maintains a redundant,
weakly-updated parameter set and infers from its diversity that it constitutes
a memory.
Similar concerns about information retention have been raised in the context of multi-modal perception~\cite{cheng2026mojito} and video generation with camera control~\cite{cheng2026moca}, where maintaining a consistent representation across time or view is crucial.
This distinction between preserving diversity and preserving informative structure is not unique to evolutionary computation. In object detection, YOLOv13 similarly argues that naive aggregation of local features fails to capture the global multi-to-multi correlations that constitute meaningful visual semantics~\cite{lei2025yolov13}.
\section{Problem Setting and Method}

\subsection{Dynamic optimization setting}
Let $\mathcal{X}=\{0,1\}^L$ be the search space and $e_t\in\mathcal{E}$ the
environment state at generation $t$, with objective $f(x;e_t)$. We use the
standard XOR-style dynamic benchmark: $e$ is a target string and
$f(x;e)=\sum_i \mathbf{1}[x_i=e_i]$. The environment changes every $T$
generations under one of two regimes:

\begin{itemize}\itemsep1pt
\item \textbf{Periodic}: $e$ alternates between two fixed states $A,B$
      differing at $d$ loci. Past states recur.
\item \textbf{Unidirectional}: each change draws a fresh uniform target. Past
      states never recur, so memory is provably useless.
\end{itemize}

The second regime is the falsification control of \S5.2, not an application
setting.

\subsection{Gated dual-copy representation}
An individual is a triple $(u,v,g)$ with $u,v\in\{0,1\}^L$ two genotype copies
and $g\in\{0,1\}^L$ a per-locus \textbf{gate} selecting which copy is read.
The evaluated phenotype is $\phi=\Gamma(u,v,g)$ for a gating rule $\Gamma$.

\paragraph{Gating rules.}
\emph{Hard gate}: $\phi_i=u_i$ if $u_i=v_i$; otherwise $\phi_i$ is the copy
selected by $g_i$. The unread copy contributes nothing to $f$, so shielding is
free. \emph{Leaky gate} (parameter $\varepsilon$): with probability
$\varepsilon$ the unselected copy is read instead, so shielding costs
$\varepsilon$ in expected evaluation accuracy.

\paragraph{Inheritance rules.}
\mix: each offspring locus independently draws one of the parent's two copies;
the gate travels with its copy. This is the standard implementation and
matches the biological analogy. \iso: copy $u$ recombines only with $u$, copy
$v$ only with $v$; the two copies form separate lineages that never exchange
material.

These two axes are the object of study. \textbf{Prior implementations fix both
simultaneously}, which is why the ablation in \S5.1 has not previously been
possible.

\subsection{Algorithm}

\begin{algorithm}[htbp]
\caption{Gated Dual-Copy EA (\textsc{gdc-ea})}
\label{alg:gdc}
\begin{algorithmic}[1]
\Require $L$, $N$, mutation rate $\mu$, gate-flip rate $\mu_g$,
         gating rule $\Gamma$, inheritance rule $\mathcal{I}$,
         truncation fraction $k$
\State initialize $u,v,g$ uniformly at random for $N$ individuals
\For{each generation $t$}
  \State $\phi \gets \Gamma(u,v,g)$
         \Comment{one copy read per locus}
  \State $\mathrm{fit} \gets f(\phi; e_t)$
  \State $S \gets$ top $kN$ individuals by $\mathrm{fit}$
  \For{each of $N$ offspring}
    \State $(p,q) \gets$ two uniform draws from $S$
    \State $(u,v,g) \gets \mathcal{I}(p,q)$
           \Comment{\mix{} or \iso}
    \State flip each locus of $u,v$ w.p. $\mu$
    \State flip each locus of $g$ w.p. $\mu_g$
           \Comment{readout}
  \EndFor
\EndFor
\State \Return best individual
\end{algorithmic}
\end{algorithm}

\paragraph{Cost.} Memory is $2\times$ the single-copy representation plus an
$L$-bit gate. Evaluation cost is $1\times$: only one copy is read per locus,
so the objective is called the same number of times as for a single-copy EA of
the same population size. All comparisons below therefore use a single-copy
baseline with \textbf{doubled population}, matching representation budget
rather than population count.

\paragraph{Degenerate cases.} With $\mu_g=0$ the hidden copy is never read and
the method reduces to a single-copy EA carrying dead weight. With the leaky
gate at $\varepsilon=1$ the two copies are always averaged and no shielding
occurs. Both appear in the ablation and both behave as predicted, which is
evidence the mechanism does what the formalization says.

\subsection{What is measured}
\paragraph{AUC.} Normalized best-individual fitness integrated over the 40
generations following each environment change. This replaces ``generations to
reach a recovery threshold,'' which is right-censored when $T$ is small and
therefore not comparable across periods.

\paragraph{Retained information (bits).} Over the $d$ loci that discriminate
the environment states, restricted to individuals whose two copies differ at
that locus, let $q_i$ be the frequency with which the hidden copy matches the
\textbf{incoming} target. Define
\begin{equation}
R=\sum_{i\in\mathcal{D}}\operatorname{sgn}\!\left(q_i-\tfrac12\right)
   \bigl[1-H_b(q_i)\bigr]\,h_i ,
\end{equation}
with $h_i$ the fraction of individuals differing at locus $i$ and $H_b$ the
binary entropy. The sign matters: a hidden copy that faithfully stores the
environment that just \emph{ended} is a negative asset, and \S5.2 shows this
case occurs. This estimator is biased under binary alphabets; see \S6.

\section{Experimental Setup}

$L=100$, $d=50$, $N=200$ with baseline $2N=400$, $\mu=0.002$, truncation
fraction $0.2$, 18 environment periods with the first 4 discarded. Ablations
use 6 seeds, main experiments 8. The baseline is a single-copy EA with
identical mutation rate, selection rule, and \textbf{recombination
operator}---see \S5.5 for why the last is not a formality.

\section{Results}

\subsection{Retention is governed by inheritance, not gating}

\begin{table}[htbp]
\centering\small
\caption{$2\times2$ ablation over the two design axes. $T{=}40$,
$\mu_g{=}0.02$, 6 seeds. Single-copy baseline AUC $=0.8715$. Retention differs
by $16\times$ across inheritance rules and barely at all across gating rules.}
\label{tab:ablation}
\begin{tabular}{llccc}
\toprule
Gating & Inheritance & AUC & Bits & Div.\ loci \\
\midrule
Hard  & \mix   & 0.8165 & $+1.28$  & 0.046 \\
Hard  & \iso   & 0.8628 & $\mathbf{+21.38}$ & 0.402 \\
Leaky & \mix   & 0.8123 & $+0.83$  & 0.040 \\
Leaky & \iso   & 0.8523 & $\mathbf{+14.76}$ & 0.325 \\
\bottomrule
\end{tabular}
\end{table}

\begin{figure}[htbp]
\centering
\includegraphics[width=\columnwidth]{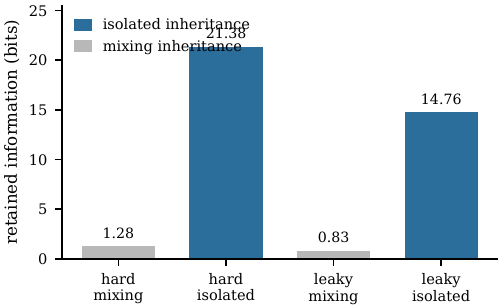}
\caption{Retained information across the $2\times2$ ablation, grouped by
inheritance rule. The gating rule---the mechanism credited in prior work---has
almost no effect.}
\label{fig:ablation}
\end{figure}

Table~\ref{tab:ablation} and Figure~\ref{fig:ablation} report the ablation.
Retention differs by $16\times$ across inheritance rules and is nearly
invariant across gating rules. AUC follows.

The last column is the diagnostic one. Under \mix, $4.6\%$ of loci still hold
differing copies---\textbf{the redundancy is intact and the diversity is
present}. What is absent is information. Shielding preserves per-locus
marginals; recombination destroys the joint pattern, and it is the joint
pattern that constitutes a memory of a past optimum.

\subsection{The memory contribution, isolated from capacity}

The two regimes are identical in representation size, population, mutation
rate, gate-flip rate, and recombination operator. They differ only in whether
past optima recur. The $0.089$ separation is therefore attributable to memory
and \textbf{not} to added capacity, implicit regularization, or increased
diversity---the three explanations a reviewer would reasonably propose, all of
which predict equal gains in both regimes.
\begin{table}[!htb]
\centering\small
\caption{AUC gain over the representation-matched single-copy baseline. Hard
gate, \iso, $\mu_g{=}0.05$, 8 seeds. The two regimes are identical in every
setting except whether past optima recur.}
\label{tab:control}
\begin{tabular}{lcc}
\toprule
$T$ & Periodic & Unidirectional \\
\midrule
40  & $\mathbf{+0.0104}$ & $\mathbf{-0.0784}$ \\
60  & $-0.0082$ & --- \\
80  & $-0.0269$ & $-0.0926$ \\
120 & $-0.0599$ & --- \\
160 & $-0.0756$ & --- \\
320 & $-0.0942$ & --- \\
\bottomrule
\end{tabular}
\end{table}

\begin{figure}[!htb]
\centering
\includegraphics[width=\columnwidth]{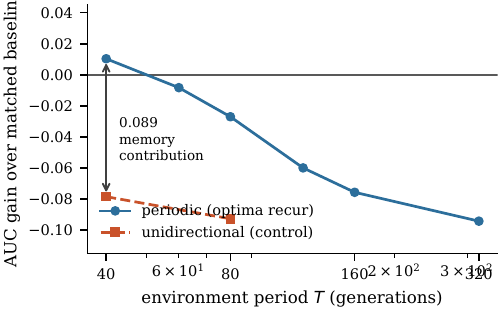}
\caption{Gain versus environment period under both regimes. The separation at
$T{=}40$ is the memory contribution; it cannot arise from added capacity,
which is identical in the two regimes.}
\label{fig:control}
\end{figure}
This yields a decomposition:
\begin{equation}
\underbrace{+0.089}_{\text{memory}}
-\underbrace{0.078}_{\text{carrying cost}}
=\underbrace{+0.010}_{\text{net}}
\end{equation}

The scientific content is the decomposition, not the net.

\paragraph{Stale memory is worse than no memory.} Gains become strongly
negative at long periods ($-0.094$ at $T{=}320$). The hidden copy, slowly
eroded across a long epoch, converges toward the environment that just
\emph{ended}, and therefore supplies a systematically wrong prior at the next
change. This is a failure mode for any implicit-memory scheme without a
refresh policy, and it is not visible in the recovery-time metrics used in
prior work.

\subsection{The readout rate is also the corruption rate}

\begin{table}[!htb]
\centering\small
\caption{Gate-flip rate sweep, two implementations differing in gating rule,
metric, and inheritance details. Both peak at $\mu_g\approx0.05$.}
\label{tab:recall}
\begin{tabular}{lccccc}
\toprule
$\mu_g$ & 0.005 & 0.02 & \textbf{0.05} & 0.10 & 0.20 \\
\midrule
Hard gate (AUC $\uparrow$)    & 0.834 & 0.863 & \textbf{0.883} & 0.870 & 0.848 \\
Leaky gate (gens $\downarrow$)& 22.5  & 17.2  & \textbf{14.3}  & 19.7  & 30.6 \\
\bottomrule
\end{tabular}
\end{table}

\begin{figure}[!htb]
\centering
\includegraphics[width=\columnwidth]{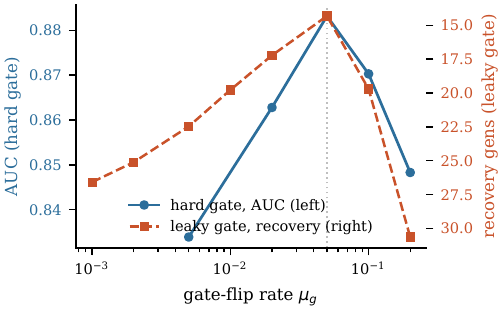}
\caption{Interior optimum in the gate-flip rate, replicated across two
implementations. Right axis is inverted so that ``up'' is better on both.}
\label{fig:recall}
\end{figure}

The gate-flip rate $\mu_g$ has two opposing effects. Too low, and the stored
content cannot be read into the evaluated phenotype. Too high, and it is
continually overwritten while the evaluated phenotype persistently carries
unselected material. An interior optimum should exist, and
Table~\ref{tab:recall} confirms one.

Retained bits peak at a lower $\mu_g$ ($\approx0.02$) than AUC does ($0.05$),
consistent with performance being approximately \emph{stored quantity $\times$
readout rate}, whose optima do not coincide.

\paragraph{Practical consequence.} $\mu_g$ is not a nuisance hyperparameter to
be tuned per problem; it is the mechanism's operating point, and it is
single-peaked, so a coarse log-scale sweep suffices.

\subsection{Negative result: gated redundancy provides no coding gain}

\begin{figure}[htbp]
\centering
\includegraphics[width=\columnwidth]{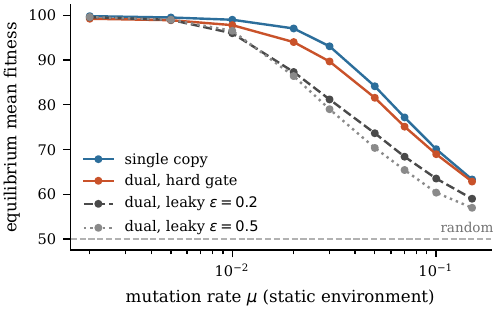}
\caption{Static environment. If redundancy provided error correction the
dual-copy arms would tolerate higher $\mu$. They do not: the threshold moves
the wrong way, and further so the leakier the gate.}
\label{fig:threshold}
\end{figure}

We sweep $\mu$ in a \textbf{static} environment and measure equilibrium
fitness. If redundancy provided error correction, the dual-copy representation
should tolerate higher $\mu$. It does not (Figure~\ref{fig:threshold}). It is
worse at every $\mu$, and worse the leakier the gate. At $\mu{=}0.02$:
single-copy $97.0$, hard-gate dual-copy $94.0$, leaky ($\varepsilon{=}0.5$)
$86.4$.

The mechanism is coding-theoretic. A repetition code improves reliability
through \textbf{joint decoding}: several copies jointly determine one output
\cite{SHANNON}. A gate reads \emph{one} copy; the other does not participate
in the decision. It is a selector, not a decoder. Adding a copy under a
selector doubles the mutational input and contributes zero decoding gain, so
it is strictly worse than a single copy in a static environment. This is
consistent with classical treatments of mutational load, in which masking
allows deleterious alleles to accumulate rather than be purged
\cite{HALDANE,MULLER}.

\textbf{This is not a tuning failure and cannot be fixed by parameter search.}
It follows from the structure of gated expression. Any method that shields
redundancy behind a selector should expect the same result.

\paragraph{Corollary.} Gated redundancy buys retention across time at the cost
of accuracy in the present. These are opposed, not two views of one benefit.

\subsection{A benchmarking hazard}

\begin{table}[htbp]
\centering\small
\caption{Baseline sensitivity to the recombination operator ($T{=}40$). The
shift is six times the effect size under study.}
\label{tab:baseline}
\begin{tabular}{lcccc}
\toprule
Baseline & 1-pt $N$ & 1-pt $2N$ & Unif.\ $N$ & Unif.\ $2N$ \\
\midrule
AUC & 0.796 & 0.810 & 0.847 & \textbf{0.872} \\
\bottomrule
\end{tabular}
\end{table}

Our own earlier runs gave the dual-copy method more effective mixing through
its slot structure while the baseline used single-point crossover on a single
copy. Changing \emph{only} the baseline's operator to uniform crossover
produced Table~\ref{tab:baseline}. The shift ($0.062$) is \textbf{six times the
effect size under study} ($0.010$). Under the weaker baseline we measured an
apparent $43\%$ improvement, nearly all of it baseline degradation.

Because a dual-copy representation changes the effective mixing pattern,
operator choice is not neutral in this comparison. We recommend that
dynamic-optimization comparisons report the recombination operator explicitly
and hold it fixed across arms.

\section{Limitations}

\paragraph{The retention estimator is biased under binary alphabets.} At a
locus where the two copies differ, the hidden copy is typically just ``the
complement of the current target,'' and with a binary alphabet ``not the
current target'' is indistinguishable from ``equal to some other target.''
Under unidirectional drift the estimator still reports $\approx+6.7$ bits where
the true value is zero. \textbf{We therefore make no quantitative claim
relating retained bits to gain.} The conclusions in \S5.2 are performance
comparisons that do not pass through the estimator. A fix requires an alphabet
of size $\geq 4$ or $\geq 3$ environment states with a test of \emph{which}
state is predicted.

\paragraph{Two environment states.} This causes the above bias and prevents
testing whether capacity scales with $\log|\mathcal{E}|$.

\paragraph{Population size unidentified.} $N$ affects both drift and selection
efficiency; the present design cannot separate them, so we report nothing on
this axis.

\paragraph{Uniform redundancy.} Every locus receives a second copy, but only
discriminative loci can hold useful memory; the rest pay cost for nothing.
This is the likely source of the $0.078$ carrying cost and the clearest route
to improving the net result.

\paragraph{Single benchmark family and scale.} All results use the XOR-style
dynamic benchmark at $L{=}100$, $N\leq400$. Generalization to Moving Peaks,
dynamic knapsack, or higher dimensions is untested.

\section{Discussion and Conclusion}

We set out to test a four-decade-old design rationale and found it
misattributed. Redundancy shielded from evaluation does not by itself retain
information about past optima; the shielded copy must additionally be
\textbf{inherited as an isolated lineage}. Under standard per-locus
inheritance the diversity survives and the memory does not.

This distinction between retained variance and retained information applies
beyond this method. Any scheme that maintains a redundant, weakly-updated
parameter set and infers from its diversity that it constitutes a memory is
subject to the same test, and the test is cheap: compare a recurring
environment against a non-recurring one under identical settings and see
whether the advantage survives.

Two further results are design-relevant. The readout rate is also the
corruption rate, so it has an interior optimum rather than a monotone
direction. And gated redundancy is a selector, not a code---it cannot be
expected to improve robustness to noise, only retention across time, and the
two trade against each other.

\paragraph{Biological note.} The mechanism was originally motivated by
recessive alleles persisting under selection on phenotypes. Our results
indicate the analogy does not transfer: sexual recombination is a \mix{}
inheritance rule, and under \mix{} the method does not retain information. The
structure that works---two non-mixing lineages, one weakly selected---resembles
clonal or reproductively isolated populations rather than diploid sexual ones.
We suggest the mechanism be named for its structure rather than its biological
inspiration.

\bibliographystyle{plain}
\bibliography{references}

\end{document}